# Integrating Large Language Models with Multimodal Virtual Reality Interfaces to Support Collaborative Human-Robot Construction Work

Somin Park[1], Carol C. Menassa[2*], Vineet R. Kamat[3]

[1] Ph.D. Candidate, Dept. of Civil and Env. Engineering, University of Michigan. E-mail: somin@umich.edu
[2*] Professor, Dept. of Civil and Env. Engineering, University of Michigan. E-mail: menassa@umich.edu
[3] Professor, Dept. of Civil and Env. Engineering, University of Michigan. E-mail: vkamat@umich.edu

**ABSTRACT**

In the construction industry, where work environments are complex, unstructured and often dangerous, the implementation of Human-Robot Collaboration (HRC) is emerging as a promising advancement. This underlines the critical need for intuitive communication interfaces that enable construction workers to collaborate seamlessly with robotic assistants. This study introduces a conversational Virtual Reality (VR) interface integrating multimodal interaction to enhance intuitive communication between construction workers and robots. By integrating voice and controller inputs with the Robot Operating System (ROS), Building Information Modeling (BIM), and a game engine featuring a chat interface powered by a Large Language Model (LLM), the proposed system enables intuitive and precise interaction within a VR setting. Evaluated by twelve construction workers through a drywall installation case study, the proposed system demonstrated its low workload and high usability with succinct command inputs. The proposed multimodal interaction system suggests that such technological integration can substantially advance the integration of robotic assistants in the construction industry.

**Keywords:** Human-Robot Collaboration (HRC); Human-Robot Interaction (HRI); Virtual Reality (VR); Multimodal interaction;

---

[*] *Corresponding author*

# 1. INTRODUCTION

In the Architecture, Engineering, and Construction (AEC) industry, the complex, unstructured and often dangerous work environments have led to the increasing interest in exploring how robots can assist humans in completing the tasks (Brosque et al. 2020, Adami et al. 2022; Park et al. 2023, Wang et al. 2023). Human-Robot Collaboration (HRC) leverages the precision, strength, and repeatability of work allowed by robotic interfaces, and blending these attributes with human workers' cognitive capability, knowledge of the craft and adaptability to change. Given that effective communication among construction workers is important for improving labor productivity (Johari and Jha 2021), sharing essential information for correct performance within human-robot teams is equally important. Members of human teams exhibit anticipatory information-sharing initiatives to accomplish collaborative tasks. To achieve this with the integration of robotic assistants into the construction industry, easy to learn and bidirectional communication between workers and robots is necessary. The importance of the user-friendly and efficient communication with the robots is highlighted by the fact that the willingness of construction workers to engage with robotic assistants is significantly affected by their perceptions of the system's ease of use and usefulness (Park et al. 2023).

Speech communication, recognized as an easy and intuitive form of human interaction, could be the fastest and most efficient way to interact with robots (Marge et al. 2022). This mode of communication has the capacity to seamlessly transmit task-related information directly without the constraints of information loss, highlighting its potential to enhance interaction with collaborative robots. In construction, this recognition of speech's utility has prompted investigations into the application of spoken or typed natural language to improve operational efficiency (Shin and Issa 2021; Linares-Garcia et al. 2022; Park et al. 2024). However, the reliance on speech inputs for task execution presents challenges, particularly in the accurate specification of necessary task attributes (Linares-Garcia et al. 2022; Park et al. 2024). Such challenges undermine the practicality and convenience of speech-based interaction, suggesting a need for further research and development.

Addressing these limitations, nonverbal cues can be leveraged with verbal communication to use the respective strengths of each mode. Specifically, gestural movements of hands offer a rich array for semantics and are easily recognizable (Yongda et al. 2018). The inclusion of the hand gestures in multimodal interfaces alongside speech introduces the utility of deictic references, such as pointing at objects during conversations. This allows for more efficient verbal communications, as it eliminates the need for detailed verbal descriptions to identify objects uniquely (Wagner et al. 2014).

For instance, instead of providing detailed descriptions like "pick up the middle block in the row of five blocks on the right" (Paul et al. 2016) or "please pick up the sheet 500300 and position it in the stud 500100" (Park et al. 2024), users can issue succinct commands like "move this to that location" or "place this one there" through pointing gestures. The confluence of speech and pointing gestures offers considerable advantages in reducing cognitive load (Goldin-Meadow et al. 2001), decreasing communication errors (Lee et al. 2013), and increasing communicative efficiency (Wagner et al. 2014). Despite these benefits, the application and potential of multimodal interfaces that combine speech and hand gestures within HRC in the construction industry remain largely unexamined.

Recognizing the underexplored potential of multimodal interfaces, particularly in the context of HRC in construction, highlights a significant opportunity for innovation in the design of the HRC interaction systems. The existing gap in effectively integrating speech and hand gestures for HRC points towards the need for comprehensive solution to implement the interaction for construction tasks. Consequently, the objectives of this study are to (1) propose a multimodal interaction method for HRC in construction; (2) devise a strategy for the integration of diverse software solutions to implement the interaction method; and (3) verify the proposed method through a user study.

To this end, this paper proposes a novel multimodal interaction system that integrates voice commands and hand controller inputs within immersive Virtual Reality (VR) environments for HRC in construction. This multimodal interaction allows users to employ a hardware controller for pinpointing workpieces on-site, while utilizing verbal commands for task specification. Moreover, it incorporates a Large Language Model (LLM) as a virtual assistant to facilitate bidirectional communication. Further enriching the system, the integration of Building Information Modeling (BIM) ensures retrieval of information about the workpieces for construction tasks. The practical application of this interaction system is evaluated by construction practitioners using a case study of a drywall installation.

## 2. RELATED WORK

Recent advancements in Natural Language Processing (NLP) have led to the application of conversational systems, using Natural Language (NL) inputs, in facilitating interaction between humans and computers, including HRC. In the construction industry, researchers have mainly explored the application of conversational systems in two domains of Information Retrieval (IR) and integration with VR/AR technologies (Saka et al. 2023). This section reviews related studies about conversational systems in construction. Following the discussion, a summarization of the limitations identified in these studies is provided.

In the construction industry, most of the efforts on conversational systems have focused on retrieving information from data sources such as BIM (Saka et al. 2023). These efforts have focused on enabling efficient retrieval of project information through query-answering systems. For instance, Lin et al. (2016) introduced a data retrieval and representation system for cloud BIM applications via text input. Further advancing the field, Zheng and Fisher (2023) developed a dialog system that not only searches BIM information through text commands but also delivers NL responses with corresponding 3D visualization.

Voice commands have also been explored to access BIM data, with studies examining their application for enhancing design and operation functionalities within BIM (Shin and Issa 2021; Elghaish et al. 2022). Shin and Issa (2021) developed a BIM Automatic Speech Recognition (BIMASR) framework to convert a BIM operating environment from expert-oriented into a user-oriented. This framework allows users to manipulate BIM data using speech commands, such as changing materials within a model. Elghaish et al. (2022) proposed a data retrieval assistant that enables BIM users to perform tasks such as the creation of a room schedule through natural language commands. The studies on information retrieval underline the progress in allowing users to efficiently interact with project information via natural language queries. However, the conversational systems rely on single-mode user inputs like text or speech, highlighting the need for the advancement of multimodal interaction systems to enhance the efficiency and intuitiveness of user interactions with other agents.

Meanwhile, by leveraging visualization functionalities of VR and Augmented Rearliy (AR), several studies have incorporated conversational systems within VR/AR in construction. In the VR domain, studies have explored the use of virtual humans to improve educational experiences in construction. Eiris-Peisari and Gheisari (2018) used a Virtual People Factory (VPF) web-based application (Rossen et la. 2009) for conversational modeling. They demonstrated its application through a high-risk hazard scenario in construction, aiming to improve student communication skills. However, this approach relied solely on text inputs from users. Wen and Gheisari (2023) focused on virtual field trips for mechanical and plumbing systems, developing a conversational system by combining VPF and Google Diagflows. This allowed users to interact with objects using computer mice and make inquiries such as "is this a hot water return pipe?" through text input. Nonetheless, the method of integrating these two types of inputs was not discussed. Both systems also have limitations due to their reliance on predefined templates for NL answers, which could restrict the flexibility and adaptability of conversations.

In the AR domain, Chen et al. (2024) addressed construction safety compliance by proposing a visual construction safety query system. This system employs a deep learning-based vision-language model, allowing users to inquire about safety issues on-site using voice input through AR glasses. Despite its advances, they analyzed spoken words alongside image scenes rather than incorporating gestures into the interaction system. Chen et al. (2020) proposed a multimodal interaction system using human hands designed for swarm robot selection that could be applied to various industrial domains including construction. This system, facilitated through AR, enables users to issue instructions like "select the robots in this range", using speech and pointing gestures. While this development provided the integration of software modules and hardware components for the use of AR in HRC, its applicability is limited to the selection of multi robots rather than facilitating collaborative tasks with the robots.

In addition to the above application areas, there are studies that explore the use of voice-based conversational systems to help workers perform construction tasks. One example is the work of Linares-Garcia (2022), who developed a voice-based intelligent virtual assistant (VIVA) specifically designed to increase the productivity of construction workers during welding tasks. The virtual agent system was developed based on Google Actions requiring expected questions and semantic knowledge (the task steps). However, users are expected to give questions such as *"After connecting PS-3 and B-8, what should I do next"* that include pieces' tag IDs or location of the items in previous steps as part of the context addition.

Ye at al. (2023) explored the impact of ChatGPT, which is one of the Large Language Models, on trust in HRC assembly task. It showed that people perceived less mental load and high trust when using a GPT-enabled robot assistant compared to using fixed control commands. Park et al. (2024) introduced a framework that enables NL-based interactions with robots for pick-and-place construction operations, demonstrating its effectiveness through drywall installation tasks. This system employs deep learning-based language models, allowing it to accurately process instructions that specify targets, destinations, and placement methods. However, the scope of these two studies was limited to a single modal interaction, necessitating precise mentions of task-relevant information, such as the names or attributes of objects, in instructions. Expanding to encompass multiple modalities in communication thus promises to improve intuitiveness and usability in interaction systems for collaboration.

All the previous studies on conversational systems in construction have primarily focused on users receiving information from virtual assistants, without addressing how users should respond when the information provided is incorrect. This oversight could be particularly significant in operation-critical environments such as construction sites, where accurate and timely information is crucial for decision-making and safety.

In the construction industry, the previous studies on NL-based conversational systems for VR/AR or robotics have the following limitations:

(1) Lack of support for HRC in Construction: Many studies have proposed interaction systems for education, safety, and other areas, rather than for robotic completion of actual construction work. There is a need for a framework that facilitates the execution of construction tasks through conversational systems in a VR environment.

(2) Limited user input channels to speech: Most of the studies depend on NL inputs, either through voice or text. To improve intuitiveness and efficiency in interaction, additional input channels can be added.

(3) Inherent limitations of conversational systems themselves: Firstly, some systems require training data to develop the conversational systems. Second, several studies rely on predefined templates for generating natural language responses. Additionally, there is research necessitating reference to previous steps in multi-step tasks. Overall, all the studies position users (or construction workers) primarily as recipients of information, where the NL answers have been designed to guide users. The intention behind the generated NL responses has been to lead users, rather than enabling users to participate as collaborative partners capable of offering feedback, providing instructions or dismissing the wrong information. To harness the potential of bidirectional communication in collaboration, the design of the conversational systems with an advanced language model is needed.

Table 1 provides a summary of these limitations. To address these challenges, this study proposes a multimodal interaction system that facilitates conversation with collaborative robots in construction. This study integrates diverse software solutions, enhancing the efficacy and scope of HRC in the construction industry.

1  Table 1. Comparison of characteristics about conversational systems to support applications on VR/AR or operations in
2  construction.

| Study | Application | AR/VR | User inputs | | Conversation | | | | Interaction command examples |
| --- | --- | --- | --- | --- | --- | --- | --- | --- | --- |
| | | | NL inputs | Others | A | B | C | D | |
| Eiris-Peisari and Gheisari (2018) | Education (hazard scenarios) | VR (desktop) | Text | N/A | Yes | Yes | N/A | Yes | Not Provided |
| Wen and Gheisari (2023) | Education (jobsite investigation) | VR (desktop) | Text | Mouse (pointing) | Yes | Yes | N/A | Yes | *Is this a hot water return pipe? / I believe the diameter is 5 inches* |
| Chen et al. (2024) | Safety Inspection | AR | Voice | Image data | Yes | No | N/A | Yes | *Are these workers safe? / What should they do to improve safety?* |
| Chen et al. (2020) | HRI | AR | Voice | Hand (pointing) | Yes | N/A | N/A | N/A | *Select the robots in this range* |
| Linares-Garcia et la. (2022) | Construction tasks (steel connection) | N/A | Voice | N/A | Yes | No | Yes | Yes | *After connecting PS-3 and B-8, what should I do next?* |
| Ye et al. (2023) | HRC | VR (headset) | Voice | N/A | No | No | N/A | Yes | *Give me a driller* |
| Park et al. (2024) | HRC | VR (desktop) | Text | N/A | Yes | N/A | N/A | Yes | *Please pick up the 4 by 8 drywall panel and hang it into the 500100 vertically* |
| Proposed system | HRC | VR (headset) | Voice | Hand controller | No | No | No | No | *Please pick up this and place it there* |

A: It requires training data to develop the interaction process.
B: It depends on predefined templates for NL answers.
C: It necessitates referencing previous steps in multi-step tasks.
D: It lacks mechanisms for human operators (or users) to verify NL answers.
N/A: Not Applicable

## 3. TECHNICAL APPROACH

### 3.1. Overview of the proposed system

Fig. 1 presents an overall framework for a multimodal interaction system in VR, designed for easy and intuitive interaction with construction robots. The proposed system requires the integration of user interaction channels, the Robot Operating System (ROS), BIM, and a game engine. The game engine, which encompasses a chat interface, enables a human operator to interact with a robot using natural language commands. Users can communicate with robots using two input channels: speech and controllers. The outputs from these channels become the input for the chat interface. The game engine plays an important role in visualizing information from BIM, displaying the construction site's data, including 3D geometric information and the semantic details of building materials. A robot engaged in interaction within the game engine is controlled via ROS, completing the system's loop of HRC.

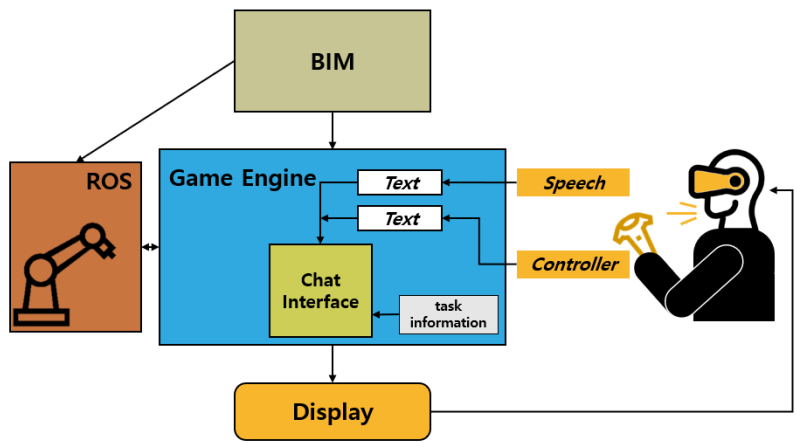

Fig. 1. Overview of the proposed system

### 3.2. Interaction interface

Fig. 2 outlines the data flow and software integration architecture for the proposed multimodal interaction system in HRC. The diagram provides a visual representation of how different components and inputs are integrated within the system. The subsequent subsections will articulate the design of the integration strategies. Section 3.2.1 will detail the methods for integrating two distinct user inputs: speech and handheld controllers captured by the VR interface. Building material information retrieved from Rhinoceros and Grasshopper is leveraged in the integration. Following that, Section 3.2.2 will concentrate on the design and flow of interaction between a human operator and a robot, specifically discussing how to implement the bidirectional communication through GPT-4 and conduct collaborative operations in ROS in the Unity game engine.

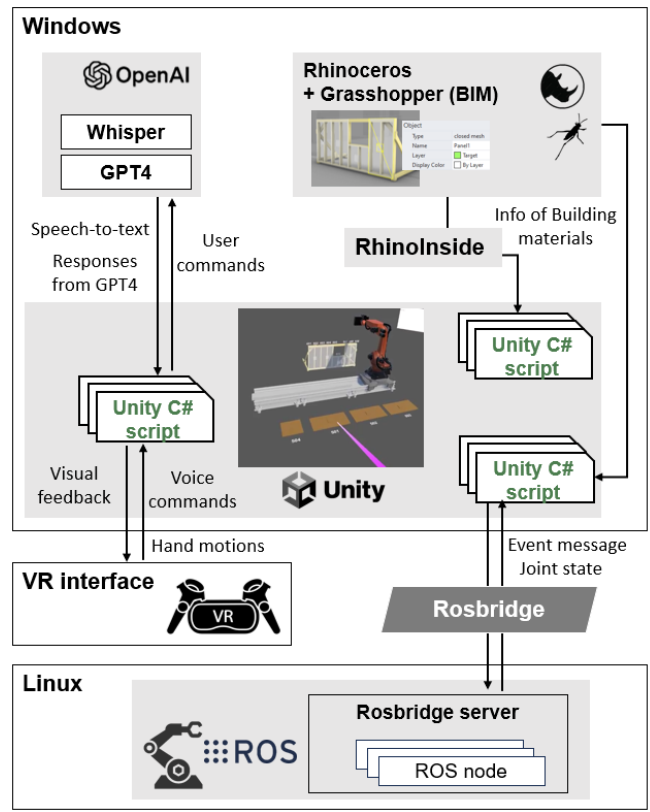

Fig. 2. Implementation of the proposed multimodal interaction system

### 3.2.1. Integration of speech and VR controller inputs

Fig. 3 shows how inputs from two different channels - speech and VR controllers – are combined in the interaction system. Voice commands are captured by the VR headset's microphone and subsequently processed by Whisper, which is an automatic speech recognition (ASR) or Speech-To-Text (STT) system (OpenAI). Whisper shows robustness in ASR since it was trained on very large datasets including 680k hours of multilingual audio data (Radford et al. 2022). While the user is engaging with the VR environment, VR controllers provide an interactive means of selecting objects within the virtual space. Selecting objects via VR controllers is streamlined using a ray interactor. The ray interactor projects a virtual beam from the controller, allowing a user to point at and select an object from a distance. To visually indicate that the object has been selected, the color of the object briefly changes to red, signaling successful engagement.

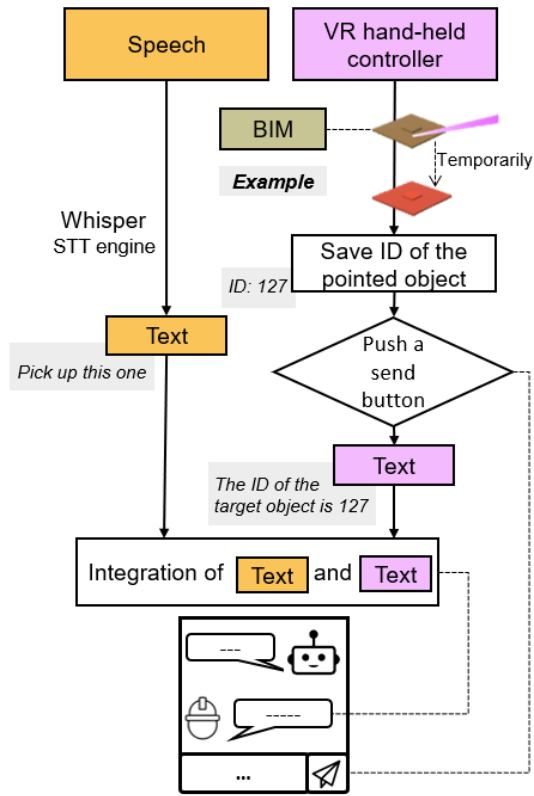

Fig. 3. Integration of inputs from voice commands and the VR controller.

A key feature of the multimodal interaction proposed in this study is the retrieval of selected object information from BIM data during the provision of user instructions. Unity generates interactable objects from BIM through the integration of Rhino and Grasshopper applications. Utilizing Rhino.Inside (McNeel 2023), an open-source add-in, Unity is enabled to concurrently operate the two applications. Once initiated alongside Unity, Grasshopper, which provides a visual programming environment, imports BIM data into Rhino. Following this, it retrieves both the geometric and semantic information of objects from the BIM data, which is subsequently transmitted to Unity.

As shown in Fig.4, the Grasshopper workspace uses multiple blocks to deliver the required BIM data in Rhinoceros to Unity for interaction purposes. This transfer results in the creation of 3D objects within Unity that are visually consistent with the Rhino model, encompassing color, geometry, and semantic information such as names, layers, and IDs. The IDs can be integrated into the user messages while the names and layers are utilized to determine the interactivity of objects within Unity. A part of C# script in Fig. 4 presents how to get various types of object data from the Rhino. While the current interaction design does not harness all the semantic data available – such as type and position, this information can be extracted and has the potential to be employed in interactions with robots.

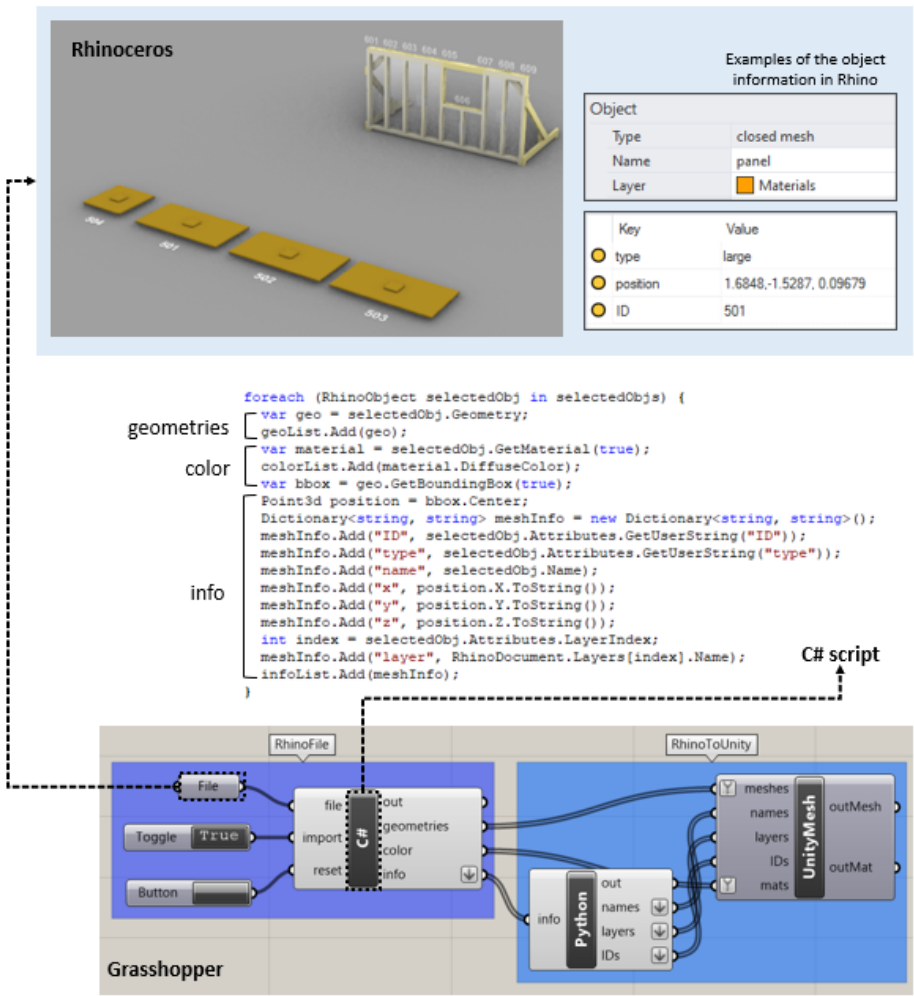

Fig. 4. BIM data in Rhinoceros and Grasshopper workspace

Upon selection of an object, its ID information is retrieved from the BIM data and stored in a text format, ready for further processing or use within the system. The retrieved ID is then held in reserve until the user activates the send button on the chat interface. For instance, the placeholder '###' in the sentence 'The ID of the target object is ###' would be replaced with the captured ID of the selected object. This textual information, representing the selected object, is then combined with the text that has been transcribed from the user's spoken command.

To illustrate, if a user verbally commands 'pick up this one' while concurrently selecting an object with the ID 127 using the controller, the consolidated command is formulated as 'pick up this one. The ID of the target object is 127.' This composite text, which contains inputs from both speech and controller commands, is sent to the chat interface when the user presses the send button of the chat interface. This process establishes a multimodal interaction framework, seamlessly integrating verbal commands with controller-based selections to enable effective communication and control within the VR space and reduces burden on user to find object IDs.

### *3.2.2. Bidirectional Communication*

In this study, the scope of the human operator's duties in the proposed interaction system includes the following activities: issuing commands to the robots, verifying the accuracy of how these commands are interpreted by the robot equipped with the capability of GPT-4, and ultimately, supervising the execution of these tasks by the robot. The conversation with a robot in the proposed interaction system is specifically designed to facilitate the first two of these activities. The necessity for the human operator to verify the interpretation of commands arises from the potential for inaccuracies due to errors in STT conversion or misinterpretations by the chat system. Therefore, the chat system is designed to detect and rectify potential errors, thereby enhancing the overall accuracy and efficiency of task execution.

Fig. 5 describes the process flowchart from the issuing the approval of the instructions. In this diagram, rectangles represent the actions executed by the human operator, while hexagons depict the robot's expected responses. Within the proposed conversation system, the user (human operator) issues instructions that include specific details about the task. The chat system then analyzes these instructions to identify the essential task information and seeks the user's confirmation. The user, serving as the ultimate decision-maker, assesses the chat system's interpretation of the task information. If the user agrees with the chat system's interpretation, they respond affirmatively, prompting the chat system to acknowledge with a reply of 'OKAY!!!'. The user then finalizes their approval by clicking an 'Approval' button on the chat interface, which triggers the robot to begin the task. Alternatively, if the user does not concur with the chat system's interpretation or if the initial instructions were incorrect, the chat system requests the user to provide the correct information, to ensure accurate task execution. This interactive process is essential for ensuring clear communication and precise task management between the human operator and the robot.

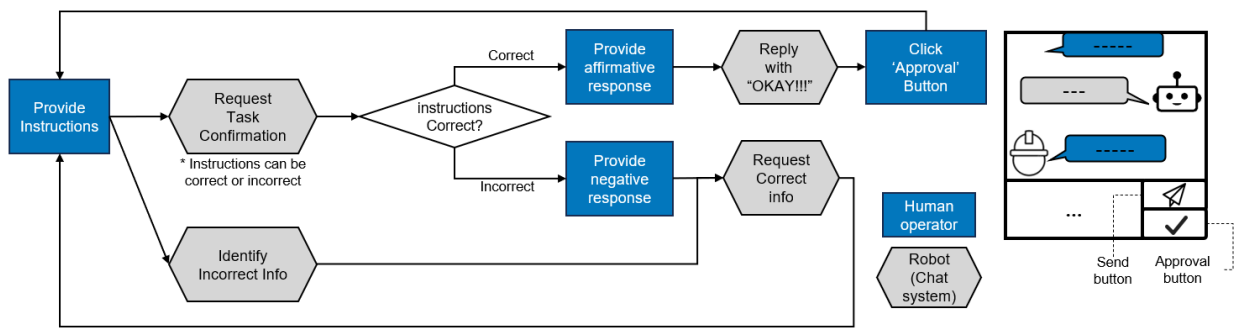

Fig. 5. Flowchart for instruction approval in the conversational system

To implement this process, GPT-4 is utilized, which is a large-sized pretrained language model that demonstrates powerful capabilities to understand and generate human language. Prompt engineering is a technique that utilizes natural language task specifications to design prompts for LLMs regarding the downstream task instead of directly altering or training the models. Without the model training with extensive data and time, this approach enables obtaining desired responses in a flexible manner and with fewer resource demands (Trad and Chehab 2024). Recent studies utilizing GPT models in construction have included elements such as task description, assistant's roles, BIM information, constraints, rules, etc. into their prompts (Ye et al. 2023; Zheng and Fisher 2023). However, these prompts, while comprehensive, are not specifically detailed or tailored for HRC as required in the proposed study. To ensure the prompt is fully aligned with the needs of the proposed interaction system, this study meticulously designs the GPT prompt with components like roles, task-specific object information, and various instructions for HRC.

The prompt of the GPT contains two contexts and four types of task instructions to build effective communication between a human operator and a robot as shown in Fig. 6. The context provided in the prompt first outlines the roles of a human operator and a robot. For example, a sentence like 'Act as a robot in the construction site and you are my teammate' in the prompt sets the tone for the text generated by the GPT. Following this context, task-related information is integrated into the prompt. This information typically includes the semantic details of construction materials, enabling the GPT to perform reasoning based on the commands issued by the human operator.

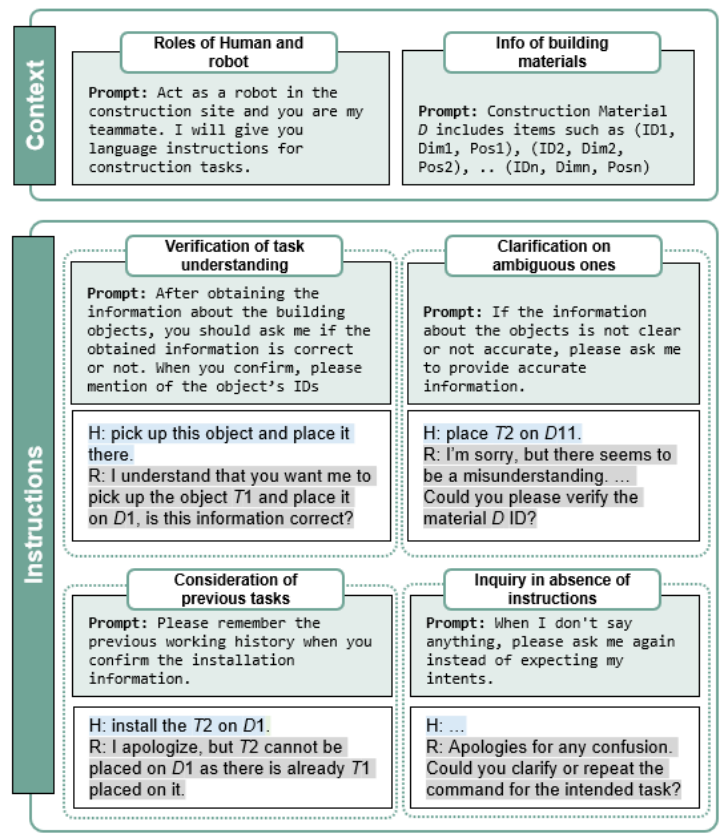

Fig. 6. Information in the prompt of GPT to manage communication scenarios with examples.

The prompt also includes four main instructions for the collaboration with robots, enabling the process flowchart described in Fig. 5. Each instruction, as shown in Fig. 6, is presented with corresponding examples showing expected robot responses (R) to human messages (H). The first instruction involves verification of task understanding, which includes GPT's interpretation of the operator's instructions and asks the operator's verification of this interpretation. This step ensures that the GPT's understanding aligns with the operator's intent. In addition, the IDs of the target objects should be mentioned when a robot asks confirmation so that the human operator can clearly identify the understanding of the robot. Second, the inclusion of clarification on ambiguous instructions is vital. This step is designed so the GPT can seek further information from the user instead of making assumptions when provided with incomplete or incorrect information.

Third, the consideration of previous tasks is integrated in the prompt. This aspect is important for multi-tasking scenarios, encouraging the robot to take into account tasks that have been previously completed. If the workpieces already used is given to the robot as task information, the robot should have ability to recognize it based on this context. The sentence in the prompt like 'please remember the previous working history when you confirm the installation information' can lead to ideal responses such as 'T2 cannot be placed on D1 as there is already T1 placed on it'. Lastly, to avoid randomly generated responses that are not related to the intent, the inclusion of inquiry in the absence of instructions is essential. This is designed to prevent the GPT from making incorrect assumptions about human intent or providing responses that are out of context when a blank message is sent. This approach ensures that even when instructions are unclear or entirely absent, the GPT is guided to respond in a manner that is more appropriate and aligned with the intended purpose.

The proposed prompt is incorporated as an essential component within the C# script for leveraging GPT functionality in Unity. On the Unity script, an API key for OpenAI is also included since it is necessary for authentication purposes in utilizing OpenAI's application. In addition, the script for GPT integration includes command lines to facilitate the integration of two input channels and selection of the GPT model. For this study, 'gpt-4,' the latest model available, was chosen to ensure the most advanced capabilities are employed. The temperature value, which is a parameter to influence the randomness in text generation, was set to 0 to minimize the variability in the responses (Zheng and Fisher 2023).

Finally, for the actualization of tasks using a robot, the Robot Operating System (ROS) running on Linux OS is employed. To do this, BIM data in Rhino and task-related information are accessed in ROS through Rosbridge using the ROS# library. As mentioned in the first instruction of the prompt, the robot is required to reference the IDs of target objects when seeking confirmation from the user. Before the user confirms by pressing an approval button, the robot's response, which includes these IDs, is processed within the Unity script. During this process, a conditional statement is used to isolate and store only the IDs in a text format in the Window OS. The stored ID information is sent to ROS once a user presses an approval button of the chat interface.

Utilizing this information, robot motion planning and robot control are conducted. The motion planning for robotic movements needs to consider collision-free motion plans designing paths where both the robot and any workpiece it carries avoid collisions with other building materials or structures in the environment. This study uses the motion planning method as proposed by Wang et al. (2021), which was developed for mobile industrial arm manipulators representing a general application for construction robotics. The calculated motion plan is then mirrored in Unity, where a virtual robot executes the movements as per the plan. The virtual robot is controlled with joint state data from ROS.

## 4. EXPERIMENTAL EVALUATION

### 4.1. Case study

An experiment that aims to assess the effectiveness of the proposed multimodal interaction system through two primary objectives was designed. The first objective is to evaluate whether the integration of speech and hand controller inputs enhances the effectiveness of HRC over a single mode of interaction. The second objective focused on assessing the effectiveness of a bidirectional interaction system, powered by a virtual assistant, in enabling precise collaboration between humans and robots to complete construction tasks. The experiment included two different interaction parts: (1) speech-based interaction and (2) multimodal interaction (speech + handheld controller) with a focus on the task of drywall installation. Twelve construction workers were recruited for the experiment and were asked to interact with a virtual robot through a VR headset, visualizing a simulated work environment as shown in Fig. 7. This environment includes a stud frame and four drywall panels, each uniquely identified by a three-digit ID.

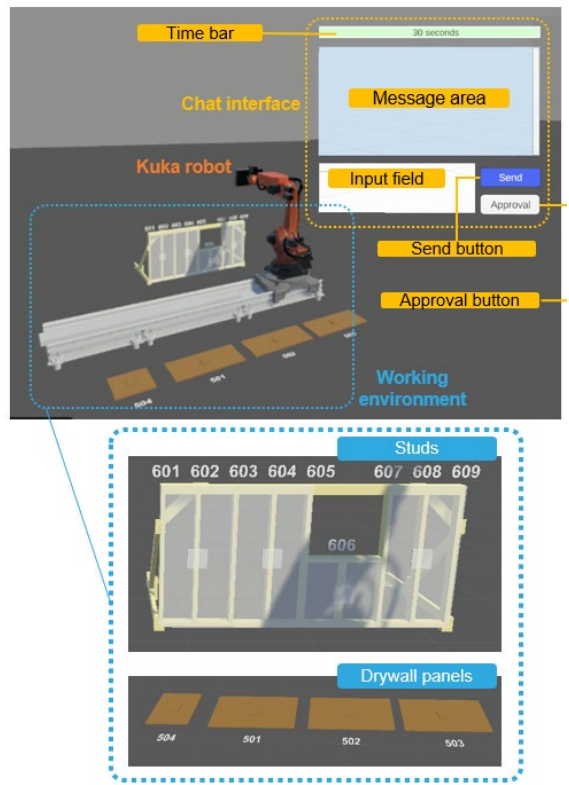

Fig. 7. VR environment in Unity

The experiment employs a 6 Degrees of Freedom (DOF) Kuka industrial robotic arm, which is mounted on a track. The scope of robot actions in this experiment is limited to 'pick-up' and 'place' activities, necessitating specific information about the target object for pick-up and the destination for placement. The experiment utilizes two sizes of panels: three standard panels measuring 4 by 8 feet and one uniquely sized panel measuring 4 by 4 feet. The panels represent target objects for pick-up while studs serve as destination for placement. The VR environment also features a chat interface consisting of several elements: an input field at the bottom, a message area in the middle, a time bar at the top, and two buttons, for sending messages and approving robot tasks, on the lower right side.

Participants are tasked with installing four panels, utilizing either speech or multimodal interactions. There is no set sequence for installing the panels. In the speech interaction, workers gave installation instructions by referring to the IDs or locations of panels and studs. For example, a worker might say, "Please place panel 504 in the second rightmost position" illustrated in Fig. 8. With multimodal interaction, workers gave instructions using demonstrative words such as "this" and "that," while pointing at objects with a handheld controller. For instance, a worker could instruct, "Please place this panel at this stud." As a result of selecting objects using the controller, sentences such as "(the ID of the target panel is 501) (the destination is the center of stud 602)" could be added in the input message. It is also permissible in multimodal interactions to mention the panel's ID while pointing at it with the controller.

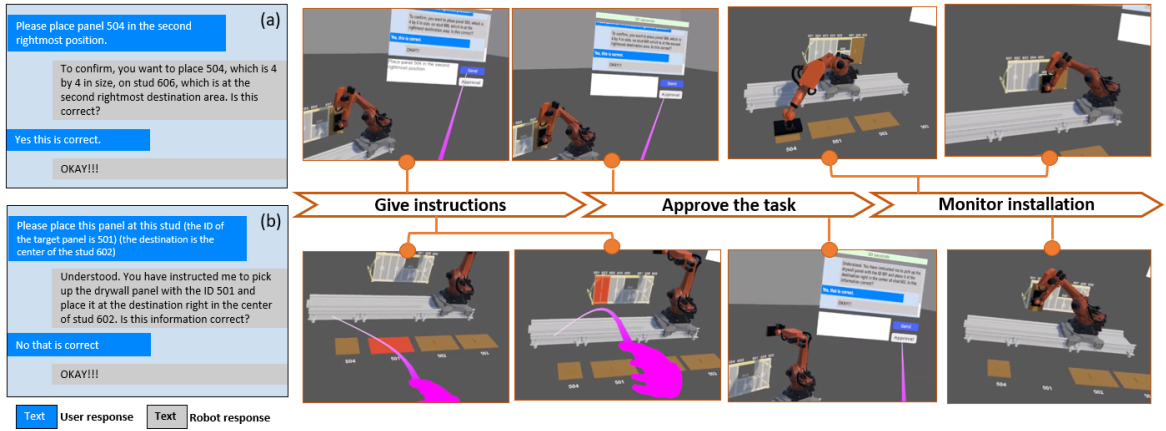

Fig. 8. Examples of HRC in VR interface: (a) Verbal interaction (b) Multimodal Interaction.

Participants are instructed to intentionally provide incorrect instructions for panel installation to evaluate the GPT model's capacity for error detection and correction within the chat application. Participants are also instructed to issue a minimum of two incorrect directives during each interaction phase. The incorrect instructions can be categorized into four types: (1) providing mismatched panel and stud pairings, (2) referring to non-existent objects, (3) instructing the placement of panels or studs that were already in place, and (4) giving a partial or duplicate information. Participants are free to randomize the introduction of these errors within their task execution.

Fig. 9 illustrates the prompts input into the GPT model for the experiment. These prompts were designed to integrate both the situational context and the specific instructions elaborated in Fig. 5, along with additional task-relevant context. This included critical specifics like the dimensions and identifiers of the target objects, as well as their intended placement locations. Notably, the automatic transfer of semantic information regarding target objects from the BIM into the GPT's prompts was not part of this study. To incorporate information about the building material into the prompts, we manually entered the information in the prompt.

**Prompt:** Act as a robot in the construction site and you are my teammate. I will give you language instructions for drywall installation. you should get the information about a target of the 'pick up' action and destination of the 'place' action. I will give you information about the target and destination. *[Roles of Human and robot]*

Targets are drywall panels. Currently, on the construction site, there are 4 drywall panels: their ID and size pairs are as follows: (501, 4 by 8), (502, 4 by 8), (503, 4 by 8), (504, 4 by 4). For example, the size of the panel 501 is 4 by 8. *[Info of building materials]*

Destinations are components of the stud frame. On the construction site, there is a stud wall. The stud wall consists of nine studs. IDs of the studs are 601,602, 603, 604, 605, 606, 607, 608, and 609. The nine studs are arranged in sequence from left to right. 601 is the leftmost stud and the 609 is the rightmost stud. There are four areas. In the center of each area, there is a destination stud. One panel will be placed on the center of the selected area. In other words, one panel will be placed on the center of the selected stud. *[Verification of task understanding]*

After obtaining the information about target and destination, you should ask me if the obtained information is correct or not and confirm the information. When you confirm, please mention the ID of the stud and panel. If my answer is 'it's correct' or 'yes', please exactly say 'OKAY!!!'.

On the center of the studs 601, 603, 605, 607 and 609, panels can't be placed. Importantly, on the center of the studs 602, 604, and 608, only 4 by 8 sized panels should be installed. Stud 602 is located on the center of the leftmost destination area. When I select the leftmost destination area, it means that the panel will be placed on the center of the stud 602. Stud 604 is located on the center of the second leftmost destination area. When I select the second leftmost destination area, it means that the panel will be placed on the center of the stud 604. Stud 608 is located on the rightmost destination area. When I select the rightmost destination area, it means that the panel will be placed on the center of the stud 608. Stud 606 is located on the second rightmost destination area.
Please remember that on the studs 606, only 4 by 4 sized panel should be installed. For example, the drywall panel 504 should be placed on the center of the stud 606 or on the second rightmost destination area. If the panel size is not corresponding to the destination information, it should not be installed. *[Contexts of the task]*

If the information about the panel or the destination is not clear or not accurate, please ask me to provide accurate information. When I give two different information about the target or destination, you should confirm which information is correct. For example, when I give two different IDs of the targets, please confirm which one is correct instead of selecting one of them yourself. *[Clarification on ambiguous ones]*

Please remember the previous working history when you confirm the installation information. For example, when I select the panel that was already installed, it cannot be installed again, and You should explain the reason why it cannot be installed. When I select the destination where a panel was already installed, it cannot be used again, and You should explain the reason why the panel cannot be installed on the destination. When you explain the reason why it cannot be installed, please also mention the available targets or destinations. *[Consideration of previous tasks]*

When I confirm the information and you say 'OKAY!!!', you should assume that the panel is installed on the stud. When you get both information about the panel and stud, you will start to install.

When I don't say anything, please ask me again instead of expecting my intents.
Let's start. Do not generate any scenarios about the situation. *[Inquiry in absence of instructions]*

Fig. 9. Prompt for GPT for drywall installation.

**4.2. Participants**

The requirement for participation in the experiment included people aged 21 or older who have prior experience working on construction sites. However, certain groups were excluded from participation to ensure safety. This exclusion applied to individuals who are pregnant, elderly, or those with pre-existing conditions that may affect their virtual reality experience, such as vision abnormalities, psychiatric disorders, or other medical conditions. Additionally, participants were required to avoid wearing glasses while using a VR headset to ensure optimal interaction and to prevent potential discomfort related to the fit of the headset.

A total of twelve construction personnel were recruited in the State of Michigan, and Table 2 shows their demographic information. The participant demographic profile indicates uniformity in gender, with the entire group consisting of male individuals. The age distribution is moderately varied, with the majority (58.33%) within the 30-39 age range, 16.67% in the 40-49 age range, and 25% being over 49 years old. Educational levels among participants show diversity: 8.33% are high school graduates or hold a General Educational Development (GED), half have some college or vocational training, 16.67% hold an associate degree, and 25% have earned a bachelor's degree.

Table 2. Demographic information of 12 participants

| Item | Characteristics | Frequency | Percentage |
|---|---|---|---|
| Gender | Male | 12 | 100.00 |
| Age | 30 – 39 | 7 | 58.33 |
|  | 40 – 49 | 2 | 16.67 |
|  | Above 49 | 3 | 25.00 |
| Education levels | High school graduate or GED | 2 | 8.33 |
|  | Some college or vocational training | 6 | 50.00 |
|  | Associate degree | 1 | 16.67 |
|  | Bachelor's degree | 3 | 25.00 |
| Job titles | Skilled Craftsmen | 2 | 16.67 |

|  | Foreman | 2 | 16.67 |
|---|---|---|---|
|  | Superintendents | 3 | 25.00 |
|  | Managers | 3 | 25.00 |
|  | Detailers | 1 | 8.33 |
|  | Instructors / Coordinators | 1 | 8.33 |
| Work experience (years) | 0 – 9 | 3 | 25.00 |
|  | 10 – 19 | 4 | 33.33 |
|  | 20 – 29 | 2 | 16.67 |
|  | Over 29 | 3 | 25.00 |

Occupational roles of the 12 participants span across the construction industry, with skilled craftsmen (Carpenter and Drywall finisher), foremen, superintendents (Piping Labor, General Trades, and General), managers (Project, BIM, and Operations) and a smaller representation from detailers, and instructors/coordinators. Despite the current job titles not necessarily involving on-site work, all 12 participants have prior experience physically working on construction sites. The range of work experience among the participants extends from under 10 years to more than 29 years, with the average work experience in the construction industry being 19.83 years for the group.

The experiment for each participant was structured into four sessions, with a total duration of 55 mins as shown in Fig. 10. During the 10-minute introduction phase, participants are briefed on what to do and shown how to use voice commands and handheld controllers to interact with a robot in a virtual environment. With the explanation of how to use a chat interface, they are also introduced to how the virtual setting visualized through a VR headset looks like.

Following the introduction, participants engaged in a 10-minute trial task, practicing with speech-based and multimodal (speech + controller) interactions. The participants were introduced to example instructions and their tasks. The main experiment, lasting 25 minutes, challenges participants to apply these interaction methods to install four drywall panels. The order in which the four panels are installed is not predetermined, allowing participants to install them in any order. Finally, participants are asked to complete a 10-minute survey on Google Forms, where they provide feedback on their experience, evaluating workload, usability, and their personal preference between the two interaction methods.

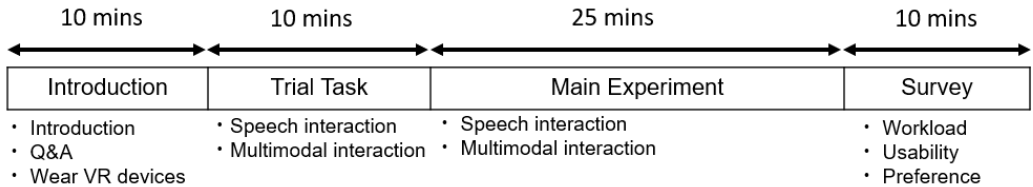

**Fig. 10**. Timeline of the experiment

## 4.3. Results and discussion

### 4.3.1. Workload

The National Aeronautics and Space Administration Task Load Index (NASA-TLX) was utilized to measure workload perceived by participants (Hart, 2006). As one of the most widely used instruments to assess overall subjective workload (Hoonakker et al. 2011; Li et al. 2019), NASA-TLX consists of six domains: Mental demand, Physical demand, Temporal demand, Performance, Effort, and Frustration. In this study, the dimension 'Effort' was divided into 'Mental effort' and 'Physical effort' to capture more understanding of the effort type. Table 3 lists the questions associated with each dimension. Participants rated each subscale on a five-point Likert scale ranging from 1 (Strongly Low) to 5 (Strongly High).

**Table 3.** NASA-TLX questionnaires

| Domain | Question |
|---|---|
| Mental demand | How much mental was required? |
| Physical demand | How much physical activity was required? |
| Temporal demand | How much time pressure did you feel due to the rate or pace at which the tasks or task elements occurred? |
| Performance | How successful do you think you were in accomplishing the goal of the task set by the experimenter (or yourself)? |
| Mental effort | How hard did you have to work (mentally) to accomplish your level of |

| | performance? |
|---|---|
| Physical effort | How hard did you have to work (physically) to accomplish your level of performance? |
| Frustration | How insecure, discouraged, irritated, stressed, and annoyed did you feel during the task? |

Fig. 11 and Fig. 12 present the assessment of the perceived workload for two interaction methods among twelve participants, using NASA-TLX scale. Fig. 11 indicates that participants' workload perception is largely consistent across most categories for both speech and multimodal interaction methods. This consistency suggests that the type of interaction method does not significantly alter the perceived workload. However, a notable variance in responses regarding mental demand is observed for speech interaction as compared to multimodal interaction, implying that speech interaction may be mentally more challenging for some participants. Additionally, one high rating in the frustration category for multimodal interaction indicates possible issues within this method that could lead to user frustration. Fig. 12 shows the mean workload scores of each dimension. The comparison between the two interaction methods shows a minimal difference, with the discrepancy in scores across all criteria being 0.5 or less. This finding highlights the relatively equivalent workload perception between speech and multimodal interactions among the participants in the experiment.

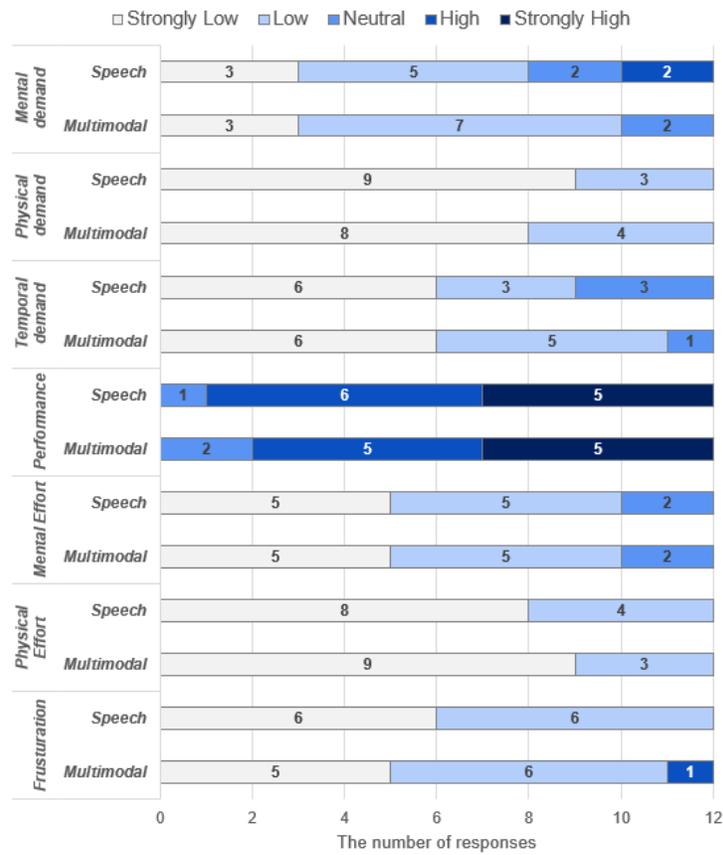

**Fig. 11.** Distribution of NASA TLX results

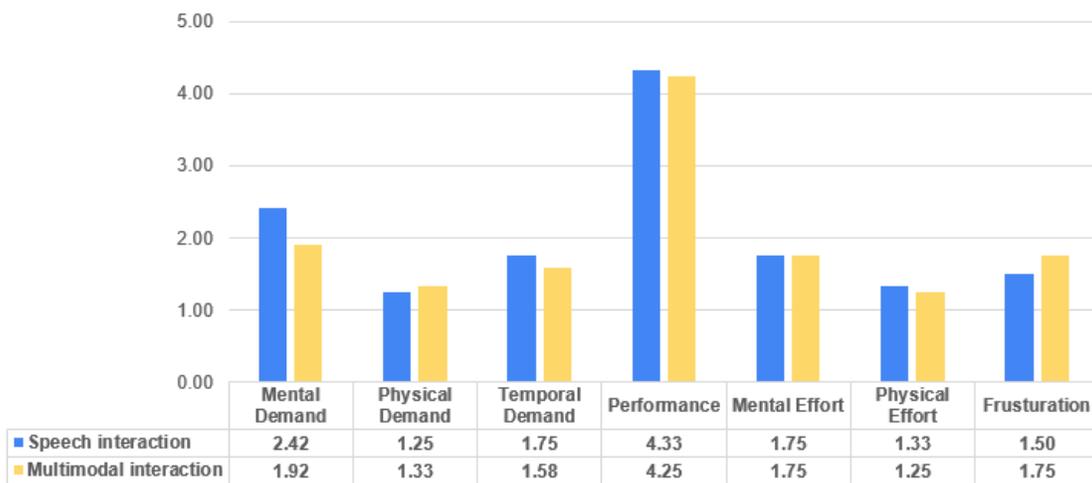

**Fig. 12.** Mean scores of NASA TLX results

*4.3.2. Usability*

To assess usability of the interaction modalities, two specific statements were used: 'This interaction method was intuitive to interact with a robot' to evaluate intuitiveness, and 'This interaction method was easy to interact with a robot' to measure ease of use. Participants expressed their level of agreement with each statement using a five-point Likert scale, ranging from 1 (Strongly Disagree) to 5 (Strongly Agree). Fig. 13 shows the average scores of the responses. While users favored speech interaction slightly more in usability, it's needed to note that both methods were perceived as relatively intuitive and easy to use, with average scores equal to or above 4 out of 5.

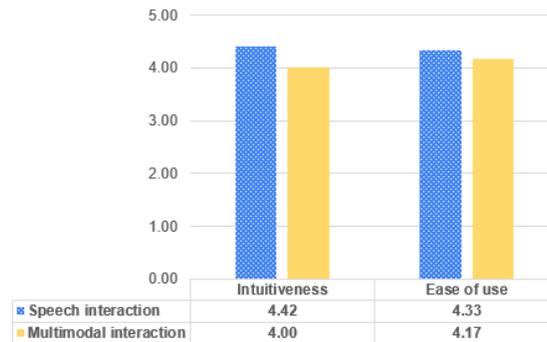

**Fig. 13.** Results of usability on interaction modalities

*4.3.3. Preferences*

To understand participants' interaction preferences within the VR environment, the survey included two questions. Initially, participants were asked 'Which type of interaction do you prefer in the VR environment?' Subsequently, to gain insight into their choices, the question 'Why do you prefer that interaction?' was posed. This led to 66.7% of participants (eight out of twelve) favoring multimodal interaction with the remaining 33.3% (four out of twelve) opting for speech interaction. Participants' justifications for their preferred interaction method varied and could be systematically categorized into five themes: efficiency, accuracy, ease of use, engagement, and versality. These categorizations are detailed in Table 4, which summarizes the participants' responses.

**Table 4.** User feedback on preferences for two interaction methods: speech and multimodal.

| Interaction method | Theme | Statements |
|---|---|---|
| Speech interaction | Efficiency | - Speech interaction seems faster.<br>- In multimodal interaction, duplicating hand gestures and some speech seems a bit redundant. |
| | Accuracy | - Speech was more accurate.<br>- I can be thorough with my intent during speech interaction. |
| Multimodal interaction | Efficiency | - It seemed quicker.<br>- It just seemed to be more efficient and quicker. |
| | Accuracy | - More accurate input, less opportunity to miss-speak or be misunderstood. |
| | Ease of use | - I feel like it is easier for me to use my words and hands at the same time when I am working.<br>- I believe that it is easier to be able to point at an object then to have the exact identity.<br>- It was an easier version of communication. I am able to just point to an object and instruct the robot to perform a task. |
| | Engagement | - I felt more involved in giving the commands to the robot |
| | Versality | - To be able to physically give instructions as well as verbal instructions. |

A notable observation is that both speech and multimodal interactions were commended for their efficiency and accuracy. Participants who preferred speech interaction highlighted its speed and precision as the reason of their choice and noted a perceived redundancy in the multimodal approach's combination of gestures and speech. In contrast, those who preferred multimodal interaction valued its higher accuracy and efficiency, noting particularly the consequent decrease in verbal miscommunication. Furthermore, multimodal interaction was also noted for its versatility, as it allows for an integration of verbal and gestural communication, which could influence the more natural and engaging way to interact with the robot. The capability to use speech and handheld controller was lauded for its ease of use, contributing to the perceived engagement and versatility of the multimodal method. These factors outline the users' preferences for their respective interaction methods in a VR environment.

*4.3.4. Performance of chat application*

Throughout the experiment, the chat application displayed commendable accuracy, generating responses that were consistent with the given instructions. During the experiment, 12 participants successfully executed four panel installations, with Fig. 14 illustrating examples of conversations in two interaction ways. There were no instances where GPT misinterpreted or responded incorrectly to correct user instructions. Notably, three participants sent blank messages to GPT by pressing the 'send' button without saying anything. In response to these instances, GPT followed the prompt for handling the absence of instructions, offering responses such as "How can I assist you further?" and "I am sorry, but I need your confirmation. Can you confirm that the information is correct?".

In analyzing the communication length between interaction methods, a notable difference was noted in the average number of words per instruction: speech interaction commands averaged 8.27 words, with the longest command "Okay, robot, could you please pick up panel 503 and install it at the rightmost portion of the framing?" reaching 19 words and the shortest being "Panel 504 to stud 606", which contains 5 words. Conversely, multimodal interaction commands were more concise, averaging 6.65 words. The longest command was "Okay, robot, could you please pick up panel 503 and install it at the rightmost portion of the framing?" comprising 18 words, while the briefest instruction "install this here," consisted of just 3 words. This distinction not only illuminates the multimodal interface's capacity to support more succinct communication but also its effectiveness in making interactions more streamlined, marking a critical enhancement in the chat interface's role in facilitating efficient and accurate HRC tasks.

Additionally, the participants intentionally issued a total of 55 incorrect instructions to test the system's capability. The chat system, powered by GPT, accurately pinpointed errors in 51 cases, achieving an accuracy rate of 92.73%. These results are described in Table 5, which details the types of incorrect instructions and the frequency with which GPT recognized and addressed the issues.

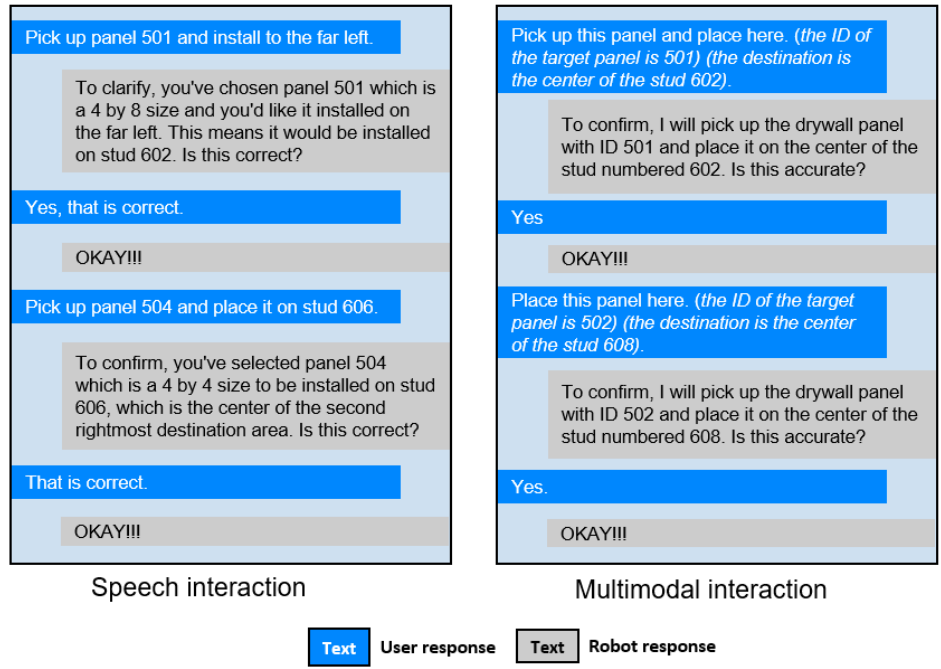

**Figure 14.** Examples of conversation for success panel installation

**Table 5.** Chat system's detection of incorrect instructions.

| Incorrect instructions | # Cases | # Issues detected |
|---|---|---|
| Mismatched pairing | 34 | 33 |
| Materials Not Present | 4 | 4 |
| Component Already Installed | 7 | 4 |
| Partial Information | 10 | 10 |
| TOTAL | 55 | 51 |

Fig. 15 further exemplifies the system's reactions to incorrect instructions across four distinct categories, with the incorrect instructions indicated by dotted lines and the inputs from handheld controllers highlighted in italics. The GPT component within the chat system showcased its analytical prowess by not only identifying wrong instructions but also by proposing alternative actions where applicable in some cases. For example, when the instruction incorrectly directed to place panel 504 on stud 605 instead of its correct paired destination, stud 606, the chat system inquired, 'would you like to install panel 504 on stud 606 instead?'

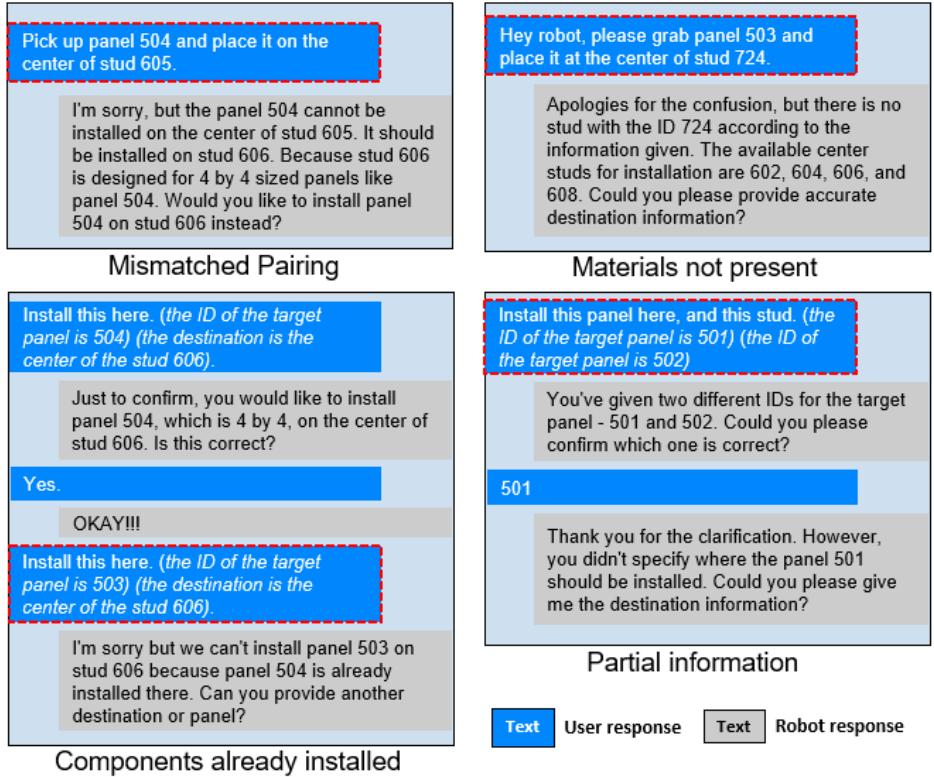

**Fig. 15.** Examples that correctly catch the incorrect instructions.

However, the system's detection was not infallible; it missed an error in one instance of mismatched pairing and in three cases where components were not reported as already installed, prompting the system to request user confirmation. The examples of these errors were shown in Fig. 16. For example, even though panel 501 cannot be installed on stud 608 because the panel 503 is already installed there, it did not catch it and it just asked if its understanding is correct. Another example shows that the chat system did not catch the fact that the panel 502 cannot be installed on the stud 606 designed for 4 by 4 sized panels like panel 504.

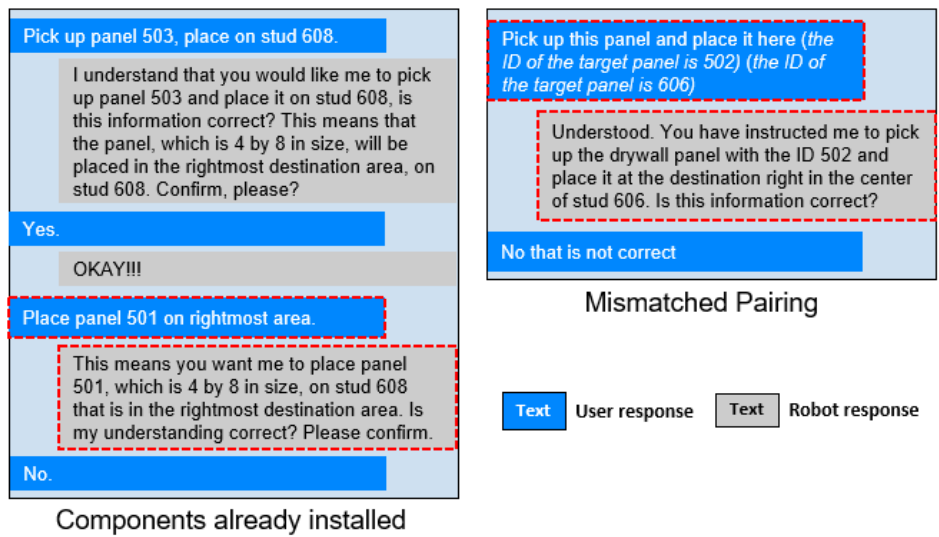

**Fig. 16.** Examples that do not catch the incorrect instructions.

*4.3.5. Discussion*

The user study conducted as part of this study has demonstrated the potential for successful deployment of the proposed multimodal interaction system within VR interfaces, integrated with a chat application, for HRC. In a case study focused on drywall installation, construction workers reported experiencing a low workload and high usability with the multimodal system. Although the assessment of workload and usability between the two interaction methods showed similar results, this uniformity may stem from the simplicity of the task, which possibly did not challenge the participants enough to discern a significant difference. This aspect suggests a need for further exploration with more complex tasks to truly gauge the differential impact on usability and workload.

However, the findings indicated a strong preference for the multimodal approach, with two-thirds of participants selecting it over speech interaction. This preference was influenced by the multimodal system's perceived efficiency, accuracy, ease of use, emergence, and versatile communication combining verbal and gestural inputs. The preference may also stem from the multimodal systems' demonstration of more concise commands on average than those of speech interaction, indicating a streamlined communication process. Efficiency and accuracy contributed to a marked preference for speech interaction, yet they were also cited as advantages of multimodal interaction. The versatile nature of multimodal interaction contributed to ease of use and helped to deepen user engagement in HRC tasks.

The GPT-4 model exhibited a high degree of responsiveness within the chat application, showing its sophisticated reasoning capabilities. Nonetheless, it did not achieve perfect accuracy. To address this, this study incorporated an additional module that allowed the GPT-4 to query its own interpretations, thereby enabling human operators to review and correct any misinterpretations. During the experiments, the chat system demonstrated an impressive accuracy rate of 92.73%, correctly identifying errors in 51 out of 55 intentional incorrect instructions. However, the system's occasional failure to detect errors in instructions underscored the need for continued human oversight.

These findings highlight the potential of integrating intuitive multimodal interfaces with Artificial Intelligence (AI)-driven chat systems in HRC in the construction industry. By reducing cognitive load and enhancing task accuracy, this innovative approach paves the way for more efficient, reliable, and user-centric HRC systems. These advancements are not only expected to improve operational efficiency in construction but also provide substantial support to construction workers in their daily tasks. Furthermore, the experimental results emphasize the critical role of a well-designed human-AI interface. While AI demonstrates its capability to reduce the cognitive load of human operators by accurately interpreting and responding to most instructions, the necessity for human oversight remains. This balanced approach could potentially lead to enhanced precision in task execution and a reduction in operational errors, fostering a more efficient and reliable HRC system in construction.

## 5. CONCLUSION

This paper proposed a multimodal interaction system for HRC in construction, leveraging VR to enhance the interaction between human workers and robots. The proposed system integrated speech and handheld controller inputs to enable easy and intuitive communication with construction robots. It employs VR controllers to point at objects of interest and NL commands to specify tasks. Furthermore, the system integrated BIM for material data retrieval, a robotic operation system for robot control, and GPT-based chat system for bidirectional communication. The practical application of the system was demonstrated through a drywall installation task, validated by twelve construction workers. Their successful completion of the task using the multimodal interaction highlighted the system's low workload and high usability.

This study makes several key contributions to the field of HRC in construction, primarily through the proposal and implementation of a multimodal interaction system. The system integrates speech and handheld controller inputs with the use of BIM data within Unity. While previous research in construction has utilized 3D game engine alongside BIM, the use of BIM data has not been previously exploited for multimodal interaction purposes. Second, a significant aspect of this study is the successful integration of diverse software components, including BIM, ROS, external servers for OpenAI API, and a game engine like Unity. This integration is critical, as it ensures that the system is not only functional but also adaptable to various construction scenarios. Additionally, the study extends domain knowledge by designing GPT prompts and proposing a conversation flow for HRC, showcasing the potential of advanced AI assistants in enhancing HRC. Next, the user study with construction workers provided in-depth qualitative and quantitative analyses on the HRC experiences, offering valuable insights into the system's operational effectiveness and user satisfaction.

However, there are some limitations to this study. Firstly, the case study focused on drywall installation, representing just one aspect of construction tasks. Future studies should expand to include a wider variety of construction scenarios and structures to fully assess the proposed multimodal interaction. Secondly, this study has the lack of automatic integration of the BIM data to the GPT prompt. This necessitated the manual inclusion of semantic information about workpieces in the GPT prompt. Automating this process in future research could significantly enhance the chat system for HRC, leading to more advanced task management and user interaction.

## ACKNOWLEDGMENTS

The work presented in this paper was supported financially by two United States National Science Foundation (NSF) Awards: 2025805 and 2128623. The support of the NSF is gratefully acknowledged.